\documentclass[11pt]{article}

\usepackage{amsmath,amssymb,bm}
\usepackage{booktabs}
\usepackage{graphicx}
\usepackage{geometry}
\usepackage{hyperref}
\usepackage{xcolor}
\usepackage{algorithm}
\usepackage{algpseudocode}

\graphicspath{{figures/}}

\title{Phys-JEPA: Physics-Informed Latent World Models for Multivariate Time-Series Forecasting}
\author{
Weizhi Nie, Weichao Liu, Honglin Guo, and Yuting Su\\
Tianjin University
}
\date{}

\begin{document}

\maketitle

\begin{abstract}
Multivariate forecasting in physical systems requires models that predict coupled temporal variables while preserving meaningful state evolution.
Deep forecasters can fit complex temporal correlations, and physics-informed models can regularize predictions with scientific constraints, but these directions are typically connected only at the decoded-output level.
Consequently, the hidden predictive state from which future trajectories are generated may be statistically useful yet physically unstructured.
We introduce Phys-JEPA, a physics-informed joint-embedding predictive architecture for multivariate time-series forecasting.
Phys-JEPA learns a latent world model in which predictive states are decomposed into physical and residual components, and physical consistency is imposed directly on latent states and latent transitions rather than only on decoded forecasts.
This formulation uses known physical variables to organize the representation space while preserving residual capacity for dynamics that are not captured by the explicit physical projection.
On Jena Climate 2009--2016, where the variables have direct meteorological meaning, Phys-JEPA reduces aggregate MSE from 0.12482 to 0.12273 and temperature MSE from 0.01892 to 0.01831 at H=24.
On Traffic, full Phys-JEPA improves aggregate MSE over the supervised baseline across all tested horizons, reducing H=192 MSE from 0.800784 to 0.773873.
On Electricity, the best variant depends on horizon: static latent consistency is strongest at H=24 and H=48, while full Phys-JEPA gives the best aggregate and target-variable MSE at H=192.
These initial results indicate that moving physics-informed learning from output space to latent predictive state space is a promising route toward interpretable and physically consistent temporal world models.
They also identify an important next step: Phys-JEPA should be studied on problems with stronger and more explicit physical constraints, where descriptor quality and transition laws can be evaluated more directly.
\end{abstract}

\section{Introduction}

Multivariate time-series forecasting is central to energy systems, transportation networks and environmental sensing.
In these domains, observations are not independent channels: electricity demand, road occupancy and other infrastructure variables are coupled through operational and physical processes.
Modern deep forecasting models can learn strong statistical predictors from historical windows, but their hidden states are usually optimized only for predictive accuracy.
As a result, a model may produce accurate short-term forecasts while learning latent dynamics that are difficult to interpret, weakly constrained by physical variables and unstable over longer horizons.

\begin{figure}[t]
    \centering
    \includegraphics[width=\linewidth]{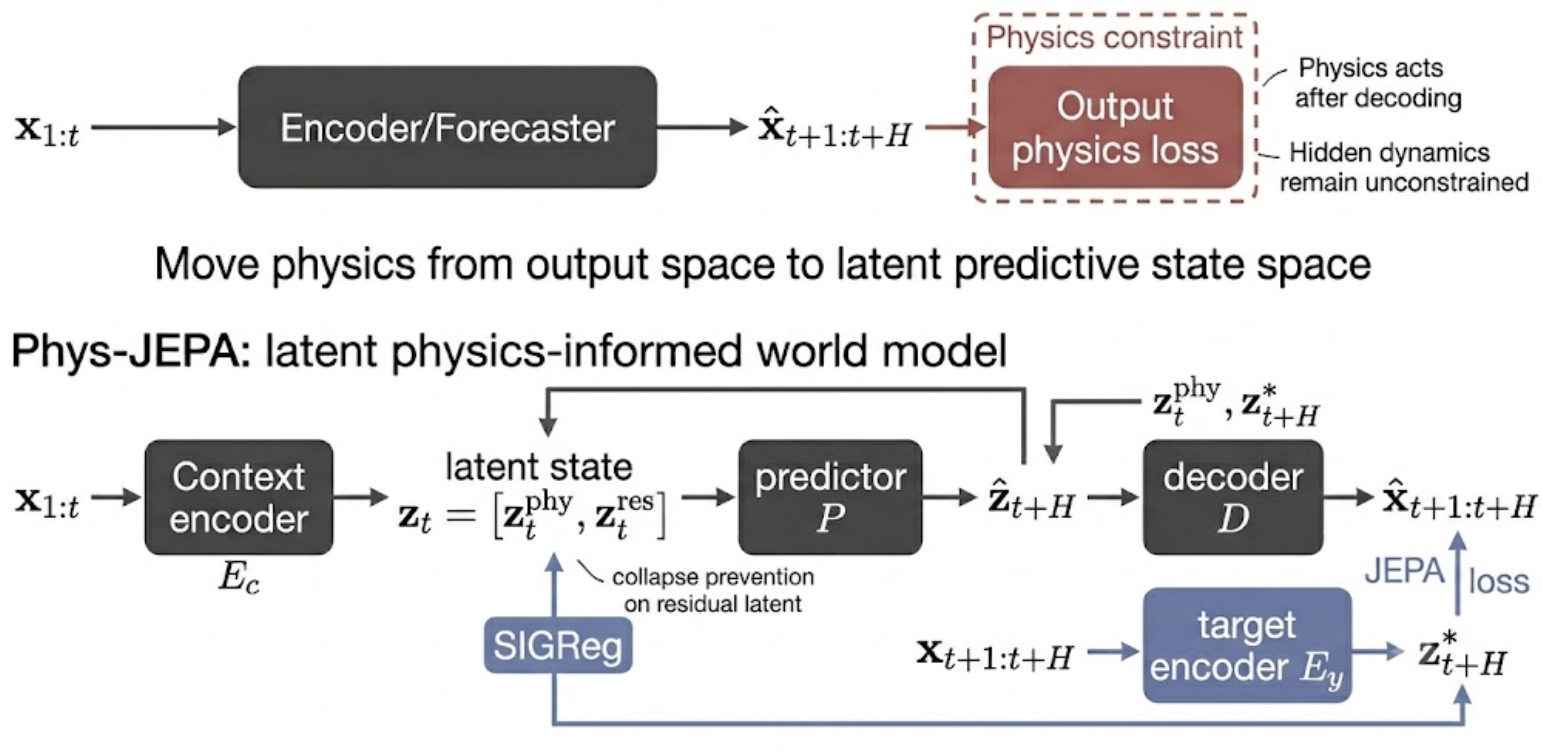}
    \caption{\textbf{Motivation of Phys-JEPA.}
    Conventional physics-informed forecasting applies physical losses after decoding, which can constrain visible forecasts while leaving hidden dynamics weakly structured.
    Phys-JEPA instead treats forecasting as latent world-model learning: the context encoder produces a decomposed latent state, the predictor forecasts the future latent state, and physical consistency is imposed directly on the latent physical component and its transition before decoding.}
    \label{fig:motivation}
\end{figure}

Physics-informed learning provides a natural way to inject scientific knowledge into neural models.
Most existing physics-informed or physics-guided forecasting methods impose constraints in the observation space, for example by penalizing decoded predictions that violate physical equations, conservation rules or algebraic relations.
This output-level strategy can improve forecast plausibility, but it leaves an important question open: if forecasts are generated from hidden predictive states, should physical consistency be enforced only after decoding, or should it shape the latent world model itself?
Figure~\ref{fig:motivation} illustrates this motivation: Phys-JEPA moves physics from decoded-output space into latent predictive state space, so that physical consistency regularizes both the learned physical state and its predicted transition before decoding.

This paper is motivated by three challenges.
\begin{itemize}
    \item \textbf{Latent states in deep forecasters are predictive but not necessarily physical.}
    Standard forecasting objectives supervise the decoded trajectory, but they do not require the hidden state to represent physically meaningful quantities or regimes.
    \item \textbf{Output-level physical regularization is indirect for long-horizon prediction.}
    Penalizing decoded forecasts can correct visible violations, yet it may not constrain the internal state transition that accumulates error across the forecasting horizon.
    \item \textbf{JEPA-style latent prediction lacks explicit physical structure.}
    Joint-embedding predictive architectures learn future representations efficiently, but their learned target space is usually defined by neural encoders rather than by known physical variables.
\end{itemize}

We propose Phys-JEPA, a physics-informed joint-embedding predictive architecture for multivariate time-series forecasting.
Phys-JEPA treats forecasting as latent world-model learning: a context encoder maps the historical window to a current latent state, a predictor forecasts the future latent state, and a decoder maps the predicted state to the future trajectory.
The key design is to decompose the latent state into a physical component and a residual component.
A physics projector reads selected physical variables from the physical component, which allows physical supervision to be applied directly to latent states.
Phys-JEPA then combines two complementary constraints: a static latent physical state loss that aligns current and future physical latent states with observed physical variables, and a dynamic latent physical transition loss that aligns the predicted physical change with the observed physical change.
The residual component is regularized separately to prevent representation collapse while preserving flexibility for non-explicit dynamics.

The resulting method moves physics-informed learning from decoded-output space to latent predictive state space.
This paper makes three contributions.
\begin{itemize}
    \item \textbf{A physics-informed latent world model for multivariate forecasting.}
    We extend JEPA-style representation prediction to physical time series by making the predictive latent state, rather than only the decoded forecast, the object of physical regularization.
    \item \textbf{A physical--residual latent decomposition with static and dynamic consistency.}
    Motivated by the need for interpretable hidden states and stable long-horizon transitions, Phys-JEPA separates physical and residual components and constrains both physical state alignment and physical transition alignment.
    \item \textbf{Empirical evidence on physical forecasting benchmarks.}
    On Traffic, full Phys-JEPA improves aggregate MSE over the supervised baseline across all tested horizons and reduces H=192 MSE from 0.800784 to 0.773873.
    On Electricity, static latent consistency is strongest at H=24 and H=48, while full Phys-JEPA gives the best aggregate and target-variable MSE at H=192.
\end{itemize}

These results provide initial evidence that physically structuring the latent predictive state is a useful direction for interpretable and physically consistent time-series world models.
They also motivate the next stage of this work: evaluation on systems with stronger explicit physical constraints, repeated-seed studies, stronger forecasting baselines and descriptor-quality ablations.

\section{Related Work}

\subsection{Long-Horizon Multivariate Time-Series Forecasting}

Long-horizon time-series forecasting has been shaped by architectures designed to capture long-range temporal dependencies, cross-variate interactions and multi-scale periodic structure.
Transformer-based forecasters first adapted attention mechanisms to long sequences by reducing the computational cost of self-attention and redesigning encoder--decoder forecasting pipelines.
Informer introduced ProbSparse attention and a generative-style decoder for long sequence forecasting \cite{zhou2021informer}, while Autoformer combined series decomposition with an auto-correlation mechanism to model periodic dependencies at the sub-series level \cite{wu2021autoformer}.
FEDformer further incorporated seasonal--trend decomposition and frequency-domain representations to improve both efficiency and accuracy \cite{zhou2022fedformer}.
These models established strong neural forecasting baselines, but they primarily treat physical variables as channels in a statistical sequence and do not explicitly constrain the learned predictive state to be physically meaningful.

Recent work has questioned whether increasingly complex attention architectures are always necessary for long-term forecasting.
The LTSF-Linear family, including DLinear, showed that simple linear decomposition models can outperform more elaborate Transformer variants on several benchmarks \cite{zeng2022dlinear}.
PatchTST revisited Transformer forecasting through patching and channel-independent weight sharing, improving long-context modeling and also supporting self-supervised pretraining \cite{nie2022patchtst}.
TimesNet transformed one-dimensional time series into two-dimensional tensors to capture intra-period and inter-period variation \cite{wu2022timesnet}, and iTransformer inverted the standard tokenization scheme so that attention operates over variate tokens rather than temporal tokens \cite{liu2023itransformer}.
These developments indicate that representation geometry, tokenization and variable interaction are central design choices in forecasting.
Phys-JEPA follows this broader direction but asks a different question: whether the predictive representation itself can be organized by physical state variables and physical state changes.

\subsection{Joint-Embedding Predictive Architectures}

Joint-embedding predictive architectures were proposed as a path toward latent-space predictive world models, in which a system predicts representations of missing or future observations rather than reconstructing the observations themselves \cite{lecun2022path}.
I-JEPA demonstrated this principle in images by predicting target-block representations from context-block representations without relying on hand-crafted augmentations or pixel-level reconstruction \cite{assran2023ijepa}.
V-JEPA extended feature prediction to video, showing that predicting latent video representations can learn visual features useful for downstream image and video tasks \cite{bardes2024vjepa}.
Related JEPA variants have explored motion-content representations, audio representations and time-series settings \cite{bardes2023mcjepa,fei2023ajepa,verdenius2024latpfn,girgis2024tsjepa,li2024tjepa}.

The JEPA family is especially relevant to physical forecasting because it separates prediction from direct signal reconstruction.
Instead of requiring the model to generate every observed detail, a JEPA model can learn a compact latent state that captures information useful for future prediction.
However, existing JEPA formulations generally define the target representation through learned encoders and regularization, not through explicit alignment with known physical variables.
Consequently, the latent space may be predictive without being physically interpretable.
Phys-JEPA extends the JEPA idea from generic latent prediction to physics-informed latent prediction by decomposing the latent state into physical and residual components and supervising the physical component through observed physical variables.

\subsection{Physics-Informed and Physics-Guided Learning}

Physics-informed neural networks incorporate scientific knowledge into neural training objectives, most prominently by penalizing residuals of governing differential equations in forward and inverse problems \cite{raissi2019pinn}.
Physics-guided neural networks and recurrent models have also integrated physical simulations, scientific constraints and physics-based losses into prediction tasks such as lake temperature modeling \cite{daw2017pgnn,jia2020physicsguided}.
These approaches show that physical knowledge can improve generalization and scientific consistency, especially when observational data are limited or when purely data-driven models produce physically implausible outputs.

Despite their importance, many physics-informed forecasting approaches impose constraints at the level of the predicted output or the explicit solution field.
For example, a decoded forecast may be penalized if it violates a conservation relation, an algebraic physical dependency or a differential-equation residual.
This output-space strategy can make predictions more physically plausible, but it does not necessarily organize the internal predictive state.
In long-horizon forecasting, where future predictions are generated from hidden states, this distinction matters: an output may be regularized after decoding while the latent dynamics remain unconstrained.
Phys-JEPA differs by applying physical supervision to latent states and latent transitions, making physical consistency part of the learned world model rather than an external correction to generated forecasts.

\subsection{Latent Dynamics and Structured State-Space Learning}

Several lines of work have studied latent dynamical systems in which hidden states evolve according to learned or physically motivated dynamics.
Neural ordinary differential equations parameterize continuous hidden-state evolution with neural networks and allow end-to-end training through ODE solvers \cite{chen2018neuralode}.
Latent ODE models extend this idea to irregularly sampled time series by combining continuous-time latent dynamics with recognition networks \cite{rubanova2019latentode}.
Hamiltonian neural networks encode conservation structure by learning dynamics consistent with Hamiltonian mechanics \cite{greydanus2019hamiltonian}, and universal differential equations combine mechanistic model components with learned neural corrections for scientific machine learning \cite{rackauckas2020ude}.
These methods motivate the broader view that forecasting models should learn structured state evolution rather than only input--output mappings.

Phys-JEPA is related to this latent-dynamics literature, but it takes a discrete-window forecasting perspective.
Instead of specifying a continuous-time differential equation for hidden-state evolution, it uses a JEPA-style predictor to map a context latent state to a future latent state.
Physical structure is introduced through a projector from the physical latent component to observed physical variables and through losses that align both latent states and latent changes with measured physical quantities.
This design keeps the training objective compatible with standard deep forecasting pipelines while adding an interpretable physical subspace to the predictive representation.

\subsection{Collapse Prevention and Latent Regularization}

Self-supervised latent prediction can suffer from representation collapse, where encoders produce uninformative constant or low-rank embeddings.
Several self-supervised methods avoid this problem through architectural asymmetry, stop-gradient operations, negative samples or explicit regularization.
BYOL showed that bootstrapped prediction with an online and target network can learn strong representations without negative pairs \cite{grill2020byol}, while SimSiam highlighted the role of stop-gradient in simple Siamese learning \cite{chen2020simsiam}.
Barlow Twins reduced redundancy by driving the cross-correlation matrix of twin-network embeddings toward the identity \cite{zbontar2021barlow}, and VICReg explicitly combined invariance, variance and covariance regularization to prevent collapse \cite{bardes2021vicreg}.

Phys-JEPA uses this line of work to stabilize temporal latent prediction, but the role of regularization must be adapted to the physical decomposition.
Applying a generic Gaussian or covariance regularizer to all latent dimensions can conflict with the structured physical component, because physical variables need not follow an isotropic latent prior.
The formulation therefore applies SIGReg primarily to the residual component, while the physical component is shaped by latent physical state and transition consistency.
This separation is intended to preserve both non-collapsed predictive representations and physically meaningful latent organization.

\section{Method}

\subsection{Overview}

Phys-JEPA is a physics-informed joint-embedding predictive architecture for multivariate time-series forecasting.
Given a context window $\bm{x}^{c}_t=\bm{x}_{t-L+1:t}$, the model predicts a future horizon $\bm{x}_{t+1:t+H}$ by forecasting a future latent state and decoding it into observations.
The central idea is to make this latent predictive state physically structured.
Instead of applying physical regularization only after the forecast is decoded, Phys-JEPA decomposes the latent state into a physical component and a residual component, and constrains both the static physical state and its dynamic transition in latent space.

\begin{figure}[t]
    \centering
    \includegraphics[width=\linewidth]{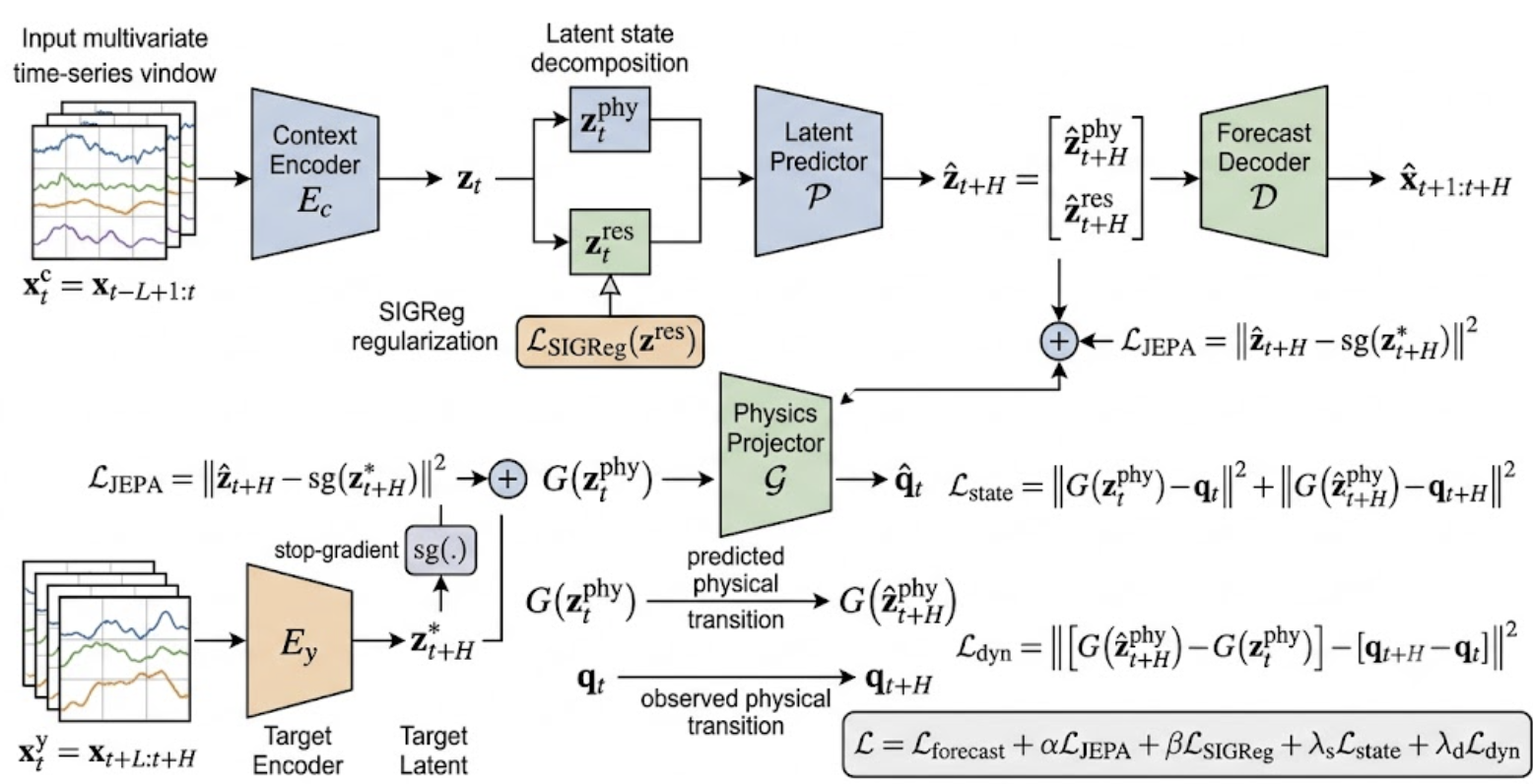}
    \caption{
    Framework of Phys-JEPA.
    A context encoder maps the historical multivariate time-series window to a latent predictive state, which is decomposed into physical and residual components.
    A latent predictor forecasts the future latent state before a decoder generates the future trajectory.
    The physical latent branch is supervised by a physics projector through both static state consistency and dynamic transition consistency, while the residual branch is regularized by SIGReg.
    }
    \label{fig:framework}
\end{figure}

Figure~\ref{fig:framework} summarizes the framework.
The upper branch follows the JEPA principle: predict the representation of the future rather than directly reconstructing the future at every intermediate step.
The lower branch injects physics into the latent world model: the physical subspace must recover selected physical variables and must evolve consistently with observed physical changes.
This design preserves the representation-learning structure of JEPA while making the predictive state interpretable and physically constrained.

\subsection{Design Rationale}

Phys-JEPA is motivated by the view that forecasting should learn a predictive state transition, not only a direct input--output mapping.
In a JEPA-style model, the predictor operates in representation space: the context representation is mapped to a future representation before any decoding step.
This makes the latent state the natural place to impose structure.
If physical constraints are applied only to $\hat{\bm{x}}_{t+1:t+H}$, they regularize the final decoded trajectory but do not necessarily shape the hidden state $\hat{\bm{z}}_{t+H}$ that produced it.
By contrast, constraining $\bm{z}^{phy}_t$ and $\hat{\bm{z}}^{phy}_{t+H}$ forces the predictive representation itself to carry physical meaning.
The physical prior therefore acts on the state used for prediction, not merely on the observable signal after prediction.

The method also distinguishes two aspects of physical consistency.
The first is \emph{state consistency}: at a given time, the physical latent state should correspond to measurable physical variables.
This is a static requirement, because it asks whether the representation has physically meaningful coordinates.
The second is \emph{transition consistency}: across a prediction interval, the physical latent state should change in a way that matches the observed physical change.
This is a dynamic requirement, because it asks whether the model has learned a meaningful evolution rule between physical states.
Using only static consistency can produce readable endpoints without explicitly constraining the displacement between them; using only dynamic consistency can align changes while leaving the absolute latent state weakly anchored.
Phys-JEPA therefore uses both constraints so that the latent trajectory is regularized in both position and motion.

This design encourages the model to learn more than a black-box map from a historical window to a future window.
A pure supervised forecaster can be written abstractly as $\hat{\bm{x}}_{t+1:t+H}=F(\bm{x}_{t-L+1:t})$, where the internal representation is not required to correspond to a state variable.
Phys-JEPA instead factorizes forecasting into a latent state, a latent transition and a decoder:
\begin{equation}
\bm{x}^{c}_t \xrightarrow{E_c} \bm{z}_t
\xrightarrow{P} \hat{\bm{z}}_{t+H}
\xrightarrow{D} \hat{\bm{x}}_{t+1:t+H}.
\end{equation}
The physical projector then tests whether the learned state and transition agree with observable physical quantities:
\begin{equation}
G(\bm{z}^{phy}_t)\approx \bm{q}_t,
\qquad
G(\hat{\bm{z}}^{phy}_{t+H})-G(\bm{z}^{phy}_t)
\approx
\bm{q}_{t+H}-\bm{q}_t.
\end{equation}
Thus, the model is rewarded not only for matching future observations, but also for organizing its predictive state so that the state can be read as physical variables and the transition can be read as physical change.
The experimental analysis tests this claim through output accuracy, ablations, horizon analysis and latent diagnostics.

\subsection{Temporal JEPA Backbone}

Let $\bm{x}_{1:T}\in\mathbb{R}^{T\times C}$ be a multivariate time series.
At time $t$, a context encoder $E_c$ maps the historical window to a latent state,
\begin{equation}
\bm{z}_t = E_c(\bm{x}^{c}_t),
\qquad
\bm{x}^{c}_t=\bm{x}_{t-L+1:t}.
\end{equation}
A target encoder $E_y$ maps the future window to a target latent representation,
\begin{equation}
\bm{z}^{*}_{t+H}=E_y(\bm{x}^{y}_t),
\qquad
\bm{x}^{y}_t=\bm{x}_{t+1:t+H}.
\end{equation}
A predictor $P$ produces a future latent state and a decoder $D$ maps it to the forecast:
\begin{equation}
\hat{\bm{z}}_{t+H}=P(\bm{z}_t),
\qquad
\hat{\bm{x}}_{t+1:t+H}=D(\hat{\bm{z}}_{t+H}).
\end{equation}

The JEPA objective aligns the predicted latent state with the target latent state using stop-gradient on the target branch:
\begin{equation}
\mathcal{L}_{JEPA}
=
\left\|
\hat{\bm{z}}_{t+H}
-
\mathrm{sg}(\bm{z}^{*}_{t+H})
\right\|_2^2.
\end{equation}
This objective encourages the context branch to learn a predictive representation of the future while avoiding a purely output-level training signal.
The decoded forecast is trained with the standard forecasting loss $\mathcal{L}_{forecast}=\|\hat{\bm{x}}_{t+1:t+H}-\bm{x}_{t+1:t+H}\|_2^2$.

\subsection{Physics-Informed Latent Decomposition}

Phys-JEPA decomposes each latent predictive state into a physical subspace and a residual subspace:
\begin{equation}
\bm{z}_t=[\bm{z}^{phy}_t,\bm{z}^{res}_t],
\qquad
\hat{\bm{z}}_{t+H}=[\hat{\bm{z}}^{phy}_{t+H},\hat{\bm{z}}^{res}_{t+H}].
\end{equation}
The physical subspace is designed to encode variables with direct physical meaning, while the residual subspace captures unresolved factors and statistical temporal variation.
For the weather setting, the physical variable vector is
\begin{equation}
\bm{q}_t =
[
p_t,
T_t,
T^{pot}_t,
rh_t,
VP^{max}_t,
VP^{act}_t,
VP^{def}_t
].
\end{equation}
We introduce a lightweight physics projector $G$ that maps the physical latent component to the physical variable space.
The projector serves two roles: it makes the physical latent state measurable, and it provides the interface through which physical constraints are applied to the latent world model.

\subsection{Static and Dynamic Latent Physics}

Phys-JEPA uses two complementary latent physics losses.
The first is a static state consistency loss, which anchors both the current and future physical latent states to observed physical variables:
\begin{equation}
\mathcal{L}_{state}
=
\left\|
G(\bm{z}^{phy}_t)-\bm{q}_t
\right\|_2^2
+
\left\|
G(\hat{\bm{z}}^{phy}_{t+H})-\bm{q}_{t+H}
\right\|_2^2.
\end{equation}
This loss implements the state-consistency principle introduced above: the physical branch should be readable as physical variables at both the source and target times.

The second is a dynamic transition consistency loss.
Static alignment alone does not explicitly constrain how physical states move across the prediction horizon.
We therefore compare the predicted physical transition in latent space with the observed physical transition:
\begin{equation}
\mathcal{L}_{dyn}
=
\left\|
\left[
G(\hat{\bm{z}}^{phy}_{t+H})-G(\bm{z}^{phy}_t)
\right]
-
\left[
\bm{q}_{t+H}-\bm{q}_t
\right]
\right\|_2^2.
\end{equation}
This loss implements the transition-consistency principle: the predicted latent transition should encode the physical change between the two states.
Together, $\mathcal{L}_{state}$ and $\mathcal{L}_{dyn}$ make the physical latent trajectory meaningful in both its endpoints and its displacement.

\subsection{Residual Regularization and Training Objective}

Because joint-embedding prediction can suffer from collapsed or redundant representations, we regularize the residual subspace with SIGReg:
\begin{equation}
\mathcal{L}_{SIGReg}
=
\|\mu(\bm{z}^{res})\|_2^2
+
\|\Sigma(\bm{z}^{res})-I\|_F^2.
\end{equation}
We apply this regularization to $\bm{z}^{res}$ rather than to the entire latent vector.
This keeps the residual branch well-conditioned while allowing the physical branch to follow the geometry induced by physical variables.

The final training objective is
\begin{equation}
\mathcal{L}
=
\mathcal{L}_{forecast}
+
\alpha\mathcal{L}_{JEPA}
+
\beta\mathcal{L}_{SIGReg}
+
\lambda_s\mathcal{L}_{state}
+
\lambda_d\mathcal{L}_{dyn}.
\end{equation}
Here $\alpha$, $\beta$, $\lambda_s$ and $\lambda_d$ balance representation prediction, residual regularization, static physical consistency and dynamic physical consistency.
Compared with output-level physics constraints, Phys-JEPA moves the physical prior inside the predictive representation: the model is trained to build a latent world state whose physical component is both readable and dynamically consistent.

\subsection{Implementation Details}

Phys-JEPA can be implemented with standard sequence encoders and multilayer-perceptron prediction heads.
In the current implementation, $E_c$ and $E_y$ encode context and target windows into fixed-dimensional latent vectors.
The predictor $P$ maps the context latent state to the future latent state, and the decoder $D$ maps the predicted future latent state to the forecasting horizon.
The latent vector is split along the feature dimension into a physical part of dimension $d_{phy}$ and a residual part of dimension $d-d_{phy}$.
The physics projector $G$ is a lightweight MLP applied only to the physical latent component.

During training, the target branch is used only to provide the JEPA target representation.
The stop-gradient operator is applied to $\bm{z}^{*}_{t+H}$ so that the predictor and context branch learn to match the future representation without collapsing the target branch into a trivial solution.
The physical losses are computed from the context physical state $\bm{z}^{phy}_t$, the predicted future physical state $\hat{\bm{z}}^{phy}_{t+H}$ and the observed physical variables $\bm{q}_t,\bm{q}_{t+H}$.
SIGReg is computed on the residual latent component, which keeps the residual space well-conditioned while avoiding an isotropic prior on the physical subspace.

At inference time, the target encoder and all loss heads are removed.
The model only uses the context encoder, latent predictor and decoder:
$\bm{x}^{c}_t \rightarrow \bm{z}_t \rightarrow \hat{\bm{z}}_{t+H} \rightarrow \hat{\bm{x}}_{t+1:t+H}$.
Thus, the physics branch introduces no additional requirement for future observations during deployment; it shapes the latent state only during training.

\begin{algorithm}[t]
\caption{Training Phys-JEPA}
\label{alg:physjepa}
\begin{algorithmic}[1]
\Require Multivariate sequence dataset $\mathcal{D}$; context length $L$; horizon $H$; physical-variable index set $\mathcal{I}_{phy}$; weights $\alpha,\beta,\lambda_s,\lambda_d$
\Require Context encoder $E_c$; target encoder $E_y$; latent predictor $P$; decoder $D$; physics projector $G$
\For{each minibatch sampled from $\mathcal{D}$}
    \State Build context window $\bm{x}^{c}_t=\bm{x}_{t-L+1:t}$ and target window $\bm{x}^{y}_t=\bm{x}_{t+1:t+H}$
    \State Extract physical variables $\bm{q}_t$ and $\bm{q}_{t+H}$ from $\mathcal{I}_{phy}$
    \State Encode current and target representations:
    $\bm{z}_t \gets E_c(\bm{x}^{c}_t)$,
    $\bm{z}^{*}_{t+H} \gets E_y(\bm{x}^{y}_t)$
    \State Predict future latent state:
    $\hat{\bm{z}}_{t+H} \gets P(\bm{z}_t)$
    \State Split latent states:
    $\bm{z}_t=[\bm{z}^{phy}_t,\bm{z}^{res}_t]$,
    $\hat{\bm{z}}_{t+H}=[\hat{\bm{z}}^{phy}_{t+H},\hat{\bm{z}}^{res}_{t+H}]$
    \State Decode forecast:
    $\hat{\bm{x}}_{t+1:t+H}\gets D(\hat{\bm{z}}_{t+H})$
    \State Compute $\mathcal{L}_{forecast}$ and $\mathcal{L}_{JEPA}=\|\hat{\bm{z}}_{t+H}-\mathrm{sg}(\bm{z}^{*}_{t+H})\|_2^2$
    \State Compute static physical loss:
    $\mathcal{L}_{state}=\|G(\bm{z}^{phy}_t)-\bm{q}_t\|_2^2+\|G(\hat{\bm{z}}^{phy}_{t+H})-\bm{q}_{t+H}\|_2^2$
    \State Compute dynamic physical loss:
    $\mathcal{L}_{dyn}=\|[G(\hat{\bm{z}}^{phy}_{t+H})-G(\bm{z}^{phy}_t)]-[\bm{q}_{t+H}-\bm{q}_t]\|_2^2$
    \State Compute residual SIGReg $\mathcal{L}_{SIGReg}(\bm{z}^{res})$
    \State Optimize
    $\mathcal{L}=\mathcal{L}_{forecast}+\alpha\mathcal{L}_{JEPA}+\beta\mathcal{L}_{SIGReg}+\lambda_s\mathcal{L}_{state}+\lambda_d\mathcal{L}_{dyn}$
\EndFor
\end{algorithmic}
\end{algorithm}

\section{Experiments}

\subsection{Datasets}

We evaluate Phys-JEPA on three multivariate forecasting benchmarks for which physical or weak-domain descriptors can be defined: Jena Climate, Electricity and Traffic.
The goal is not to benchmark a generic JEPA forecaster on arbitrary time-series datasets, but to test whether physical regularization is more useful when applied to latent predictive states.
Datasets without an explicit physical descriptor, such as ETTh1 and ETTm1 in the available result set, are therefore excluded from the main manuscript experiments and retained only as internal backbone sanity checks.

\paragraph{Jena Climate.}
The Jena Climate 2009--2016 dataset contains meteorological observations recorded at the weather station of the Max Planck Institute for Biogeochemistry in Jena, Germany \cite{jenaClimate2009}.
It provides physically meaningful atmospheric variables including pressure, temperature, potential temperature, dew-point temperature, humidity, vapor-pressure terms, air density, wind speed and wind direction.
Compared with Electricity and Traffic, Jena Climate is the most direct physical benchmark in this paper because its observed channels have explicit meteorological interpretations.
We use the cleaned multivariate version from the existing result files and set temperature, $T$ (degC), as the target variable.
The physical descriptor used by Phys-JEPA follows the method section and includes pressure, temperature, potential temperature, relative humidity and vapor-pressure variables.

\paragraph{Electricity.}
The Electricity benchmark contains electricity consumption time series from many clients.
It originates from public electric-load data distributed through the UCI Machine Learning Repository \cite{dua2019uci,trindade2015electricity} and has become a standard multivariate long-term forecasting benchmark through LSTNet and later LTSF studies \cite{lai2018lstnet,zhou2021informer,wu2021autoformer,nie2022patchtst}.
It represents a grid-load forecasting setting in which temporal behavior reflects demand cycles, aggregate usage patterns and external factors.
In our experiments, Electricity is treated as a weak-domain physical benchmark: the descriptor used for physics regularization is not an exact governing equation, but a domain signal intended to provide coarse physical or operational structure, following the broader idea of using scientific or domain constraints to guide neural forecasting models \cite{daw2017pgnn,jia2020physicsguided,raissi2019pinn}.
This distinction is important because the transition constraint may become overly restrictive when the physical descriptor is weak or highly aggregated.

\paragraph{Traffic.}
The Traffic benchmark contains road occupancy measurements from a large sensor network.
The data source is related to the California freeway Performance Measurement System (PeMS), a large-scale loop-detector traffic archive used in transportation research \cite{bickel2008measuringtraffic}, and the processed multivariate benchmark has been widely reused in LSTNet-style and Transformer-style forecasting evaluations \cite{lai2018lstnet,zhou2021informer,wu2021autoformer,nie2022patchtst}.
Traffic has strong temporal dynamics and spatially coupled demand patterns.
As with Electricity, the physical signal used here is a weak-domain descriptor rather than an exact law.
The dataset is therefore useful for testing whether latent physical state and transition consistency can improve forecasting when the variables describe an infrastructure process with meaningful temporal evolution.

\begin{table}[t]
\centering
\caption{Datasets used in the main Phys-JEPA experiments. Jena Climate provides directly interpretable meteorological variables, while Electricity and Traffic use weak-domain physical descriptors.}
\label{tab:datasets}
\resizebox{\linewidth}{!}{%
\begin{tabular}{llll}
\toprule
Dataset & Domain & Sampling / scale & Role in this paper \\
\midrule
Jena Climate & Meteorology & Weather station variables & Physical weather benchmark \\
Electricity & Grid load & Multi-client demand & Weak-domain physical benchmark \\
Traffic & Road occupancy & Sensor-network occupancy & Weak-domain physical benchmark \\
\bottomrule
\end{tabular}
}
\end{table}

\subsection{Protocol and Metrics}

All experiments use context length $L=96$, following common long-horizon forecasting protocols used by Transformer-style, decomposition-based, linear and patch-based forecasting models \cite{zhou2021informer,wu2021autoformer,zhou2022fedformer,zeng2022dlinear,nie2022patchtst,liu2023itransformer}.
We evaluate horizons $H\in\{24,48,96,192\}$ for the horizon study and focus on $H=24$ for the main ablation analysis, matching the short-to-long horizon range commonly reported in long-term forecasting benchmarks \cite{zhou2021informer,wu2021autoformer,zhou2022fedformer,zeng2022dlinear,nie2022patchtst,wu2022timesnet,liu2023itransformer}.
We report mean squared error (MSE) and mean absolute error (MAE), the standard metrics used by the main LTSF baselines considered in this paper \cite{zhou2021informer,wu2021autoformer,zhou2022fedformer,zeng2022dlinear,nie2022patchtst,wu2022timesnet,liu2023itransformer}, computed both over all variables and on a designated target variable.
For Jena Climate, the target variable is temperature; for Electricity and Traffic, the target variables are the selected benchmark channels reported in the result files.

The result set is single-seed.
We therefore interpret small differences conservatively and emphasize consistent patterns across datasets, horizons and ablations rather than isolated numerical wins.

\subsection{Compared Methods}

We compare the following model families.
\begin{itemize}
    \item \textbf{Baseline}: a supervised forecasting model trained with the decoded forecast loss, following the standard direct supervised forecasting setup used in long-horizon forecasting comparisons \cite{zhou2021informer,wu2021autoformer,zhou2022fedformer,zeng2022dlinear,nie2022patchtst}.
    \item \textbf{JEPA}: the Temporal JEPA backbone with latent prediction and Gaussian SIGReg, adapting the joint-embedding predictive principle to temporal forecasting \cite{lecun2022path,assran2023ijepa,bardes2024vjepa,girgis2024tsjepa,verdenius2024latpfn}.
    \item \textbf{Phys-JEPA}: the proposed model with both static latent state consistency and dynamic latent transition consistency, designed to combine JEPA-style latent prediction with physics-guided learning objectives \cite{lecun2022path,raissi2019pinn,daw2017pgnn,jia2020physicsguided}.
\end{itemize}

We additionally evaluate ablations to isolate the source of improvement:
\begin{itemize}
    \item output-level physical regularization, following the common physics-informed strategy of penalizing decoded predictions or output-space residuals \cite{raissi2019pinn,daw2017pgnn,jia2020physicsguided};
    \item \textbf{Static-only}: Phys-JEPA with only latent physical state consistency;
    \item \textbf{Dynamic-only}: Phys-JEPA with only latent physical transition consistency;
    \item \textbf{Static+Dynamic}: the full Phys-JEPA objective with both latent physical losses;
    \item JEPA variants without SIGReg and with GMM-SIGReg. These ablations are motivated by the role of variance and covariance regularization in preventing representation collapse in self-supervised learning \cite{zbontar2021barlow,bardes2021vicreg}.
\end{itemize}

\subsection{Implementation Setting}

Unless otherwise specified, JEPA-based models use latent dimension 128, hidden dimension 128, batch size 256 and Adam optimization.
For the best available Phys-JEPA configurations, the physical latent dimension is set to 32.
The main latent-physics runs use a JEPA weight of 0.02 and SIGReg weight of 0.001.
The static and dynamic physical weights are selected from the available sweeps.
When long-horizon runs are available only for the static loss, we explicitly refer to the model as \textbf{Phys-JEPA-Static} rather than the full Phys-JEPA.

All source result tables are stored in \texttt{analysis\_tables/}, and all plotted source data are stored in \texttt{figures/source\_data/}.

\section{Results and Analysis}

\subsection{Main H=24 Comparison}

\begin{figure}[t]
    \centering
    \includegraphics[width=0.92\linewidth]{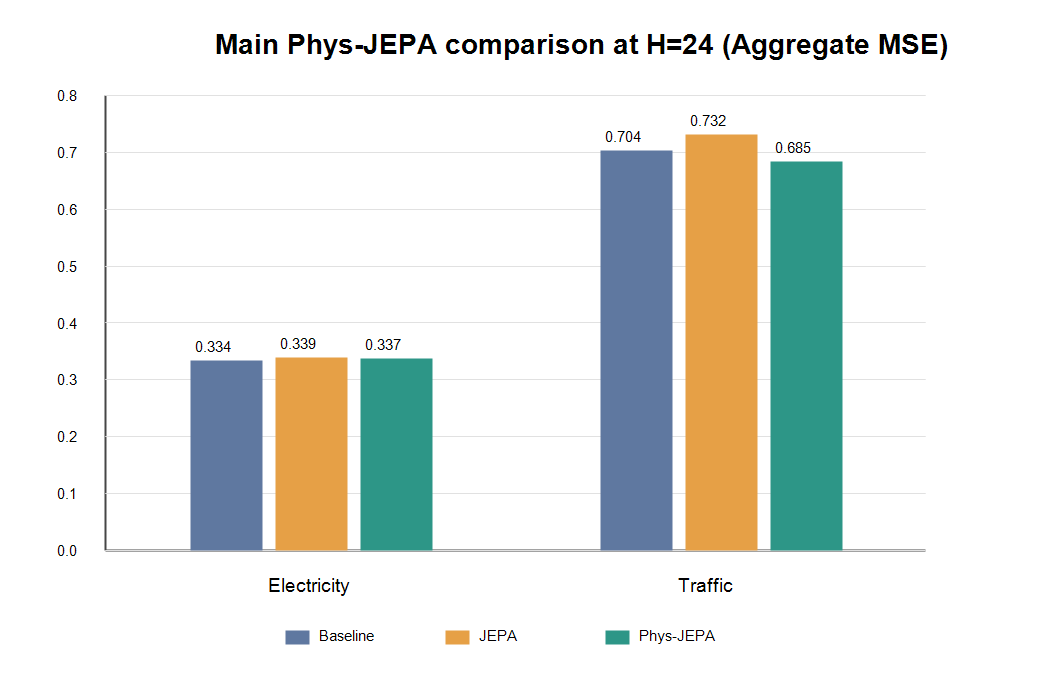}
\caption{
Core H=24 aggregate MSE comparison on Electricity and Traffic.
These are the infrastructure datasets where weak-domain physical descriptors were defined; Jena Climate is reported separately as a weather case study.
}
    \label{fig:core-h24}
\end{figure}

Figure~\ref{fig:core-h24} summarizes the H=24 aggregate MSE on Electricity and Traffic, and Table~\ref{tab:main-h24} additionally reports the Jena Climate weather case.
The comparison focuses on Baseline, JEPA and Phys-JEPA because the central question is whether latent physical constraints improve a JEPA-style predictive representation \cite{lecun2022path,assran2023ijepa}.

\begin{table}[t]
\centering
\caption{Main H=24 results on physics-relevant benchmarks. Lower is better.}
\label{tab:main-h24}
\begin{tabular}{llrrrr}
\toprule
Dataset & Method & All MSE & All MAE & Target MSE & Target MAE \\
\midrule
Jena Climate & Baseline & 0.124820 & 0.194220 & 0.018920 & 0.096560 \\
Jena Climate & JEPA & 0.124740 & 0.194220 & 0.018600 & 0.095930 \\
Jena Climate & Phys-JEPA & \textbf{0.122730} & \textbf{0.191490} & \textbf{0.018310} & \textbf{0.094500} \\
Electricity & Baseline & 0.334035 & 0.395620 & 0.329102 & 0.439064 \\
Electricity & JEPA & 0.339003 & 0.399651 & 0.340414 & 0.448069 \\
Electricity & Phys-JEPA & 0.336799 & 0.398513 & 0.354296 & 0.452776 \\
Traffic & Baseline & 0.703518 & 0.382845 & 0.536363 & 0.522612 \\
Traffic & JEPA & 0.731792 & 0.406476 & 0.557872 & 0.539948 \\
Traffic & Phys-JEPA & \textbf{0.684625} & \textbf{0.371513} & \textbf{0.506928} & \textbf{0.502751} \\
\bottomrule
\end{tabular}
\end{table}

The behavior differs between Electricity and Traffic.
On Electricity, the full Phys-JEPA objective does not improve over the baseline at H=24, while the Static-only ablation gives the best aggregate result.
On Traffic, full Phys-JEPA gives the strongest result, reducing aggregate MSE from 0.703518 to 0.684625 and target MSE from 0.536363 to 0.506928.
This supports the central hypothesis in a qualified form: latent physical constraints are most useful when the physical descriptor captures meaningful temporal evolution, whereas dynamic transition consistency can over-constrain the model when the descriptor is coarse or weakly specified.

\subsection{Jena Climate Weather Case Study}

Jena Climate provides a more physically interpretable setting than the infrastructure benchmarks because its observed variables are meteorological state variables rather than weak aggregate descriptors.
Table~\ref{tab:jena-weather} reports the previous H=24 weather results on the cleaned Jena Climate 2009--2016 split.
Phys-JEPA reduces aggregate MSE from 0.12482 to 0.12273 and temperature MSE from 0.01892 to 0.01831.
The static-only variant gives the best aggregate MSE in this run, while the full static+dynamic Phys-JEPA objective gives the best temperature MAE and a strong temperature MSE.
This is consistent with the central design claim: physical losses are most reliable when the latent physical branch is anchored to variables with direct physical meaning.

\begin{table}[t]
\centering
\caption{Jena Climate H=24 weather forecasting results. Target metrics are computed on temperature $T$ (degC). Lower is better.}
\label{tab:jena-weather}
\begin{tabular}{lrrrr}
\toprule
Method & All MSE & All MAE & Temperature MSE & Temperature MAE \\
\midrule
Baseline & 0.12482 & 0.19422 & 0.01892 & 0.09656 \\
JEPA & 0.12474 & 0.19422 & 0.01860 & 0.09593 \\
Phys-JEPA-Static & \textbf{0.12261} & \textbf{0.19126} & 0.01849 & 0.09508 \\
Phys-JEPA & 0.12273 & 0.19149 & \textbf{0.01831} & \textbf{0.09450} \\
\bottomrule
\end{tabular}
\end{table}

\begin{figure}[t]
    \centering
    \includegraphics[width=0.92\linewidth]{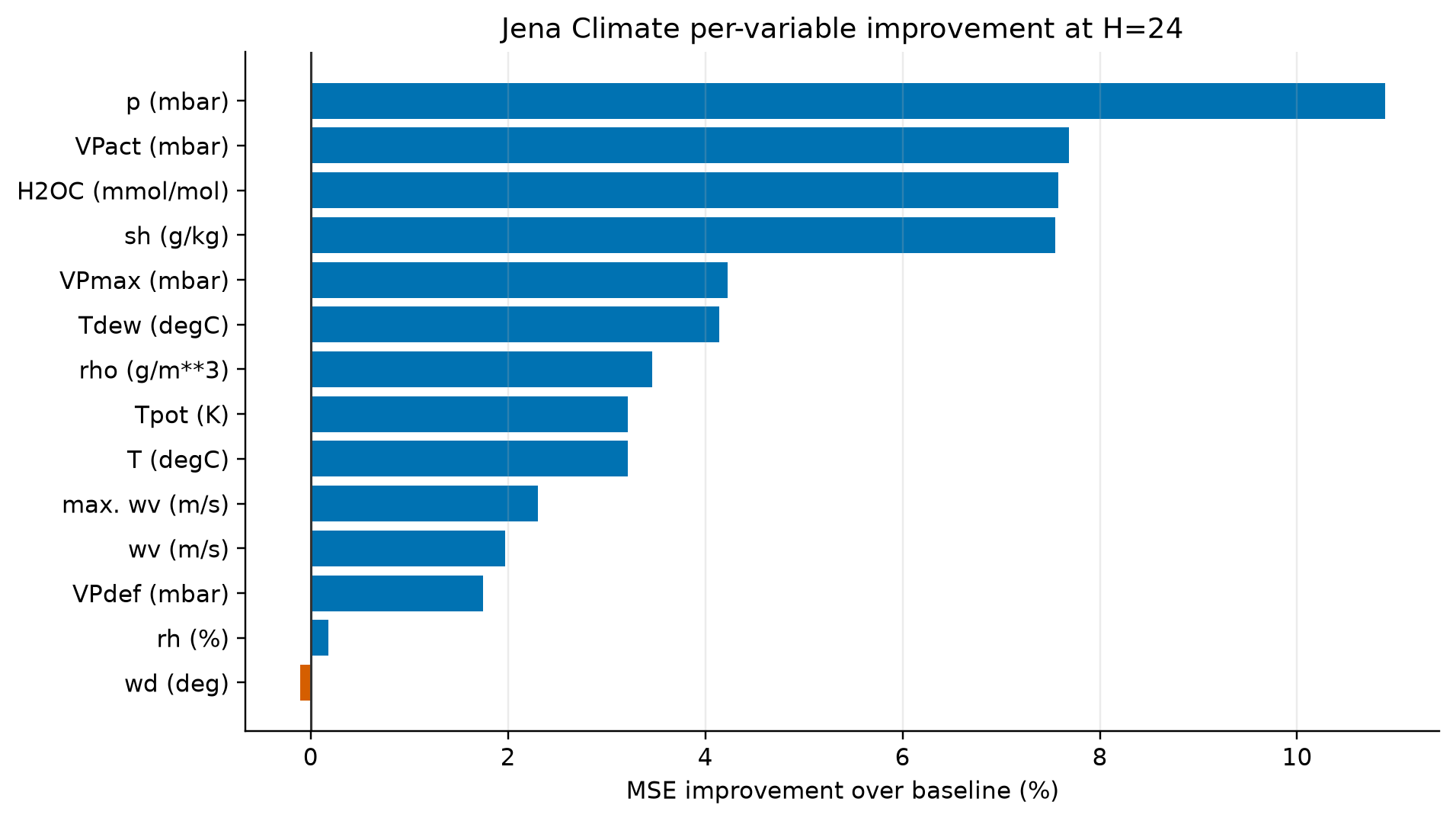}
    \caption{
    Per-variable MSE improvement of Phys-JEPA over the supervised baseline on Jena Climate at H=24.
    Phys-JEPA improves most thermodynamic and vapor-pressure variables, while wind direction remains difficult because it is angular and was treated as a scalar degree variable in the previous result set.
    }
    \label{fig:jena-per-feature}
\end{figure}

Figure~\ref{fig:jena-per-feature} gives a feature-level view of the same result.
Phys-JEPA improves pressure, temperature, potential temperature, dew-point temperature, vapor pressure, specific humidity, water-vapor concentration, air density and wind-speed variables.
The exception is wind direction, where the scalar degree representation is suboptimal because angular variables are periodic.
This identifies a concrete follow-up for the weather benchmark: encode wind direction using sine and cosine components before applying physical latent alignment.

\subsection{Horizon Analysis on Physics-Relevant Benchmarks}

\begin{figure}[t]
    \centering
    \includegraphics[width=0.92\linewidth]{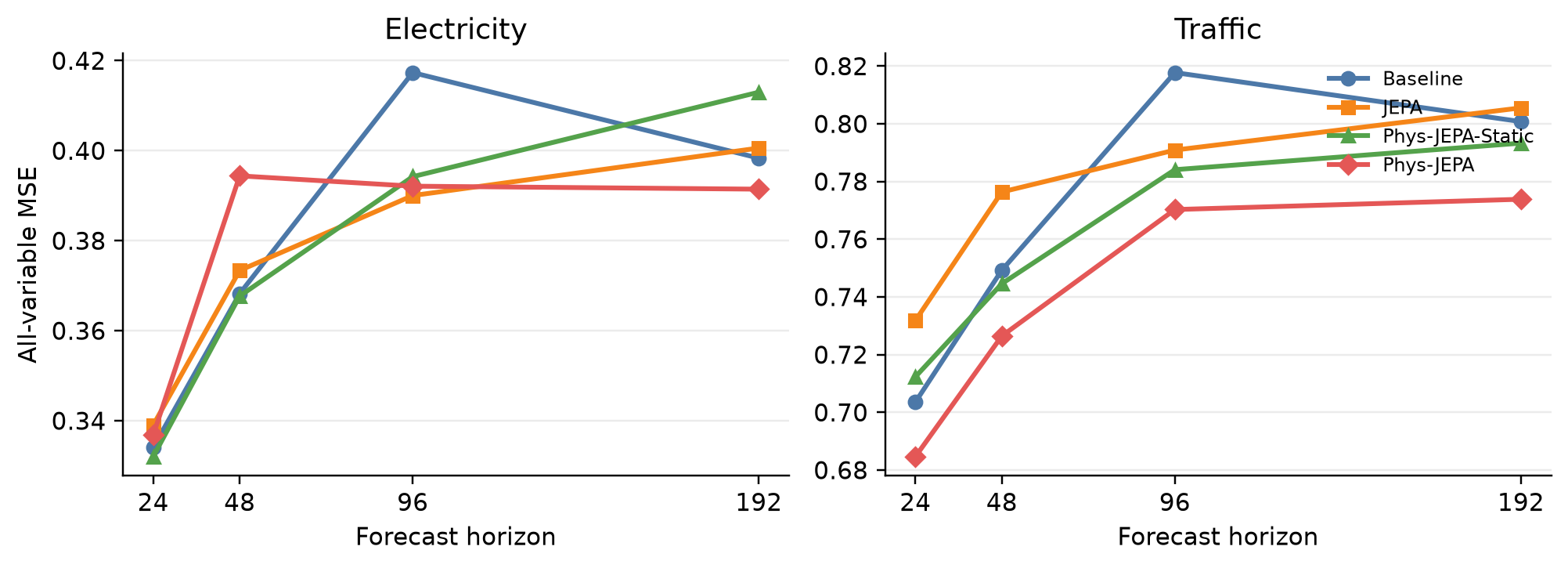}
    \caption{
    Completed horizon analysis on Electricity and Traffic.
    Full Phys-JEPA consistently improves Traffic across all horizons and gives the best Electricity result at H=192.
    }
    \label{fig:horizon}
\end{figure}

Figure~\ref{fig:horizon} reports the completed horizon sweep on Electricity and Traffic.
The comparison includes the supervised baseline, the JEPA backbone, Phys-JEPA-Static and the full Phys-JEPA objective.

\begin{table}[t]
\centering
\caption{Best method by dataset and horizon on physics-relevant benchmarks. Lower MSE is better.}
\label{tab:best-horizon}
\begin{tabular}{lllrlr}
\toprule
Dataset & Horizon & Best All method & All MSE & Best target method & Target MSE \\
\midrule
Electricity & 24 & Phys-JEPA-Static & 0.331997 & Baseline & 0.329102 \\
Electricity & 48 & Phys-JEPA-Static & 0.367690 & Phys-JEPA-Static & 0.444332 \\
Electricity & 96 & JEPA & 0.390032 & Phys-JEPA & 0.483457 \\
Electricity & 192 & Phys-JEPA & 0.391452 & Phys-JEPA & 0.501240 \\
Traffic & 24 & Phys-JEPA & 0.684625 & Phys-JEPA & 0.506928 \\
Traffic & 48 & Phys-JEPA & 0.726473 & Phys-JEPA & 0.526501 \\
Traffic & 96 & Phys-JEPA & 0.770301 & Phys-JEPA & 0.558793 \\
Traffic & 192 & Phys-JEPA & 0.773873 & Phys-JEPA & 0.595946 \\
\bottomrule
\end{tabular}
\end{table}

\begin{table}[t]
\centering
\caption{Full horizon results for Phys-JEPA and comparison methods. Lower is better.}
\label{tab:full-horizon}
\begin{tabular}{llrrrr}
\toprule
Dataset & Method & H=24 & H=48 & H=96 & H=192 \\
\midrule
Electricity & Baseline & 0.334035 & 0.368238 & 0.417358 & 0.398348 \\
Electricity & JEPA & 0.339003 & 0.373387 & \textbf{0.390032} & 0.400588 \\
Electricity & Phys-JEPA-Static & \textbf{0.331997} & \textbf{0.367690} & 0.394231 & 0.413001 \\
Electricity & Phys-JEPA & 0.336799 & 0.394410 & 0.392119 & \textbf{0.391452} \\
Traffic & Baseline & 0.703518 & 0.749287 & 0.817774 & 0.800784 \\
Traffic & JEPA & 0.731792 & 0.776401 & 0.790915 & 0.805546 \\
Traffic & Phys-JEPA-Static & 0.712296 & 0.744604 & 0.784122 & 0.793326 \\
Traffic & Phys-JEPA & \textbf{0.684625} & \textbf{0.726473} & \textbf{0.770301} & \textbf{0.773873} \\
\bottomrule
\end{tabular}
\end{table}

Traffic shows the clearest horizon robustness.
Full Phys-JEPA is the best method on Traffic for all tested horizons, reducing aggregate MSE from 0.749287 to 0.726473 at H=48, from 0.817774 to 0.770301 at H=96 and from 0.800784 to 0.773873 at H=192 relative to the supervised baseline.
It also improves the target-variable MSE at all Traffic horizons.
Electricity is more mixed.
Static consistency is strongest at H=24 and H=48, JEPA gives the best aggregate MSE at H=96, and full Phys-JEPA gives the best aggregate and target MSE at H=192.
This pattern supports the design motivation for using both static and dynamic constraints: the static loss stabilizes state alignment, while the dynamic loss becomes more useful when long-range evolution must be predicted.
The weaker Electricity gains at short horizons also suggest that dynamic consistency can be sensitive to descriptor quality in highly aggregated demand data.

\subsection{Ablation of Physical Constraints}

\begin{figure}[t]
    \centering
    \includegraphics[width=0.88\linewidth]{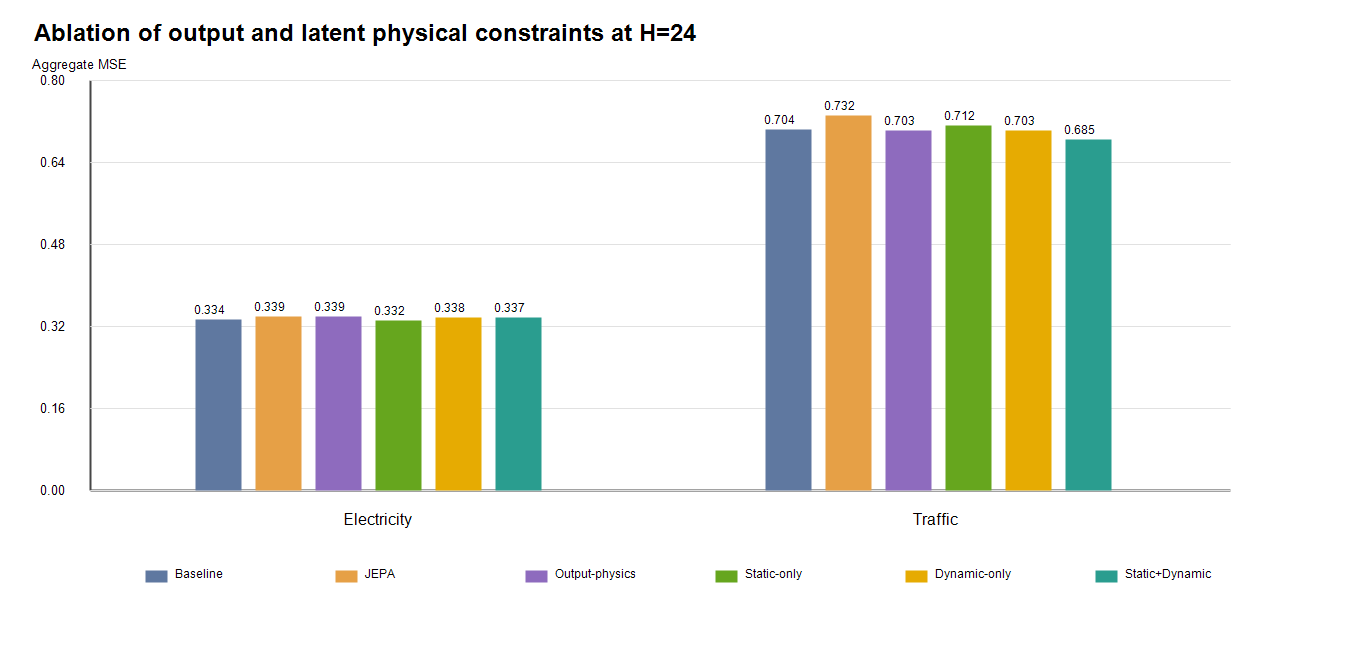}
    \caption{
    Ablation of output-level and latent physical constraints at H=24 on Electricity and Traffic.
    Static-only uses only latent physical state consistency, Dynamic-only uses only latent physical transition consistency, and Static+Dynamic is the full Phys-JEPA objective.
    }
    \label{fig:ablation}
\end{figure}

Figure~\ref{fig:ablation} separates the effects of output-level physics, static latent physics, dynamic latent physics and their combination at H=24.

\begin{table}[t]
\centering
\caption{Ablation of physical constraints at H=24. Lower aggregate MSE is better. Static-only and Dynamic-only are ablations of the two Phys-JEPA losses; Static+Dynamic is the full Phys-JEPA objective.}
\label{tab:ablation-h24}
\begin{tabular}{llrr}
\toprule
Dataset & Method & All MSE & Relative change vs. baseline \\
\midrule
Electricity & Baseline & 0.334035 & 0.00\% \\
Electricity & JEPA & 0.339003 & +1.49\% \\
Electricity & Output-physics & 0.338992 & +1.48\% \\
Electricity & Static-only & \textbf{0.331997} & \textbf{-0.61\%} \\
Electricity & Dynamic-only & 0.338176 & +1.24\% \\
Electricity & Static+Dynamic & 0.336799 & +0.83\% \\
Traffic & Baseline & 0.703518 & 0.00\% \\
Traffic & JEPA & 0.731792 & +4.02\% \\
Traffic & Output-physics & 0.703055 & -0.07\% \\
Traffic & Static-only & 0.712296 & +1.25\% \\
Traffic & Dynamic-only & 0.702893 & -0.09\% \\
Traffic & Static+Dynamic & \textbf{0.684625} & \textbf{-2.69\%} \\
\bottomrule
\end{tabular}
\end{table}

The ablation supports two conclusions.
First, output-level physics is weak in the current setting.
On Electricity, output physics increases aggregate MSE from 0.334035 to 0.338992; on Traffic, it gives only a negligible improvement from 0.703518 to 0.703055.
This supports the argument that physical regularization should act on the latent predictive state rather than only on decoded forecasts.

Second, static and dynamic latent constraints play different roles.
Electricity favors Static-only, which reduces aggregate MSE to 0.331997, while the full Static+Dynamic objective is worse than the baseline.
Traffic favors the full Static+Dynamic objective, which reduces aggregate MSE by 2.69\% and target MSE by 5.49\% relative to the baseline.
This pattern suggests that transition consistency is beneficial when the descriptor captures meaningful evolution, but can be harmful when the descriptor is too coarse to define a reliable transition rule.

\subsection{Latent Representation Diagnostics}

\begin{figure}[t]
    \centering
    \includegraphics[width=0.88\linewidth]{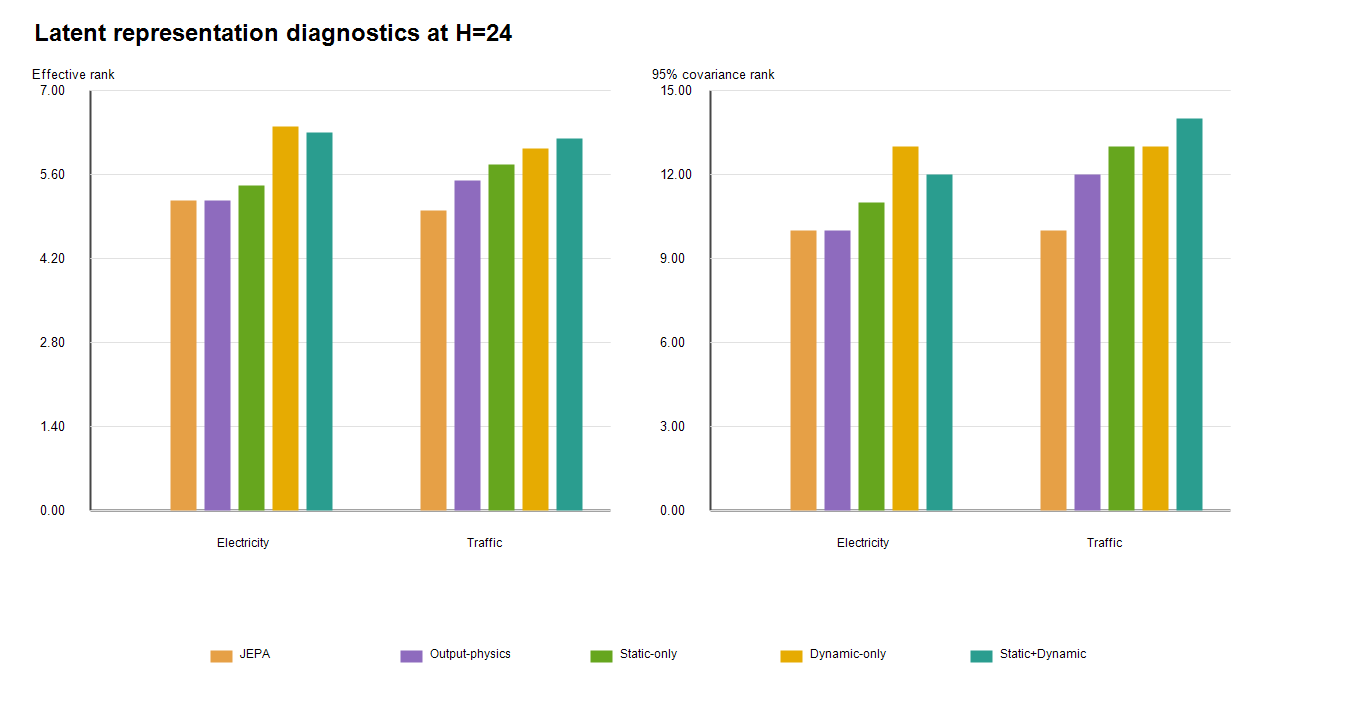}
    \caption{
    Latent representation diagnostics at H=24.
    Latent physics variants generally increase effective rank or covariance rank on Electricity and Traffic, indicating that the physical losses alter representation geometry rather than acting only as output penalties.
    }
    \label{fig:latent-diagnostics}
\end{figure}

A central claim of Phys-JEPA is that latent physical constraints change the representation space rather than merely changing the decoded output.
Figure~\ref{fig:latent-diagnostics} examines this using effective rank and 95\% covariance rank at H=24.

\begin{table}[t]
\centering
\caption{Latent representation diagnostics at H=24. The increase in effective rank and covariance rank indicates that latent physical losses change representation geometry.}
\label{tab:latent-diagnostics}
\begin{tabular}{llrrr}
\toprule
Dataset & Method & Effective rank & Cov. rank 95\% & Mean latent variance \\
\midrule
Electricity & JEPA & 5.159 & 10 & 0.399 \\
Electricity & Output-physics & 5.159 & 10 & 0.400 \\
Electricity & Static-only & 5.422 & 11 & 0.403 \\
Electricity & Dynamic-only & \textbf{6.400} & \textbf{13} & 0.371 \\
Electricity & Static+Dynamic & 6.292 & 12 & 0.355 \\
Traffic & JEPA & 4.997 & 10 & 0.426 \\
Traffic & Output-physics & 5.508 & 12 & 0.443 \\
Traffic & Static-only & 5.775 & 13 & 0.450 \\
Traffic & Dynamic-only & 6.028 & 13 & 0.443 \\
Traffic & Static+Dynamic & \textbf{6.194} & \textbf{14} & 0.443 \\
\bottomrule
\end{tabular}
\end{table}

On Electricity, JEPA has effective rank 5.16 and covariance rank 10.
Static-only increases these values to 5.42 and 11, Dynamic-only to 6.40 and 13, and the full Static+Dynamic objective to 6.29 and 12.
On Traffic, JEPA has effective rank 5.00 and covariance rank 10, while the full Phys-JEPA objective increases them to 6.19 and 14.
These diagnostics support the method claim that latent physics acts on the predictive representation itself.
They do not by themselves prove physical causality, but they show that the proposed losses reshape the latent state space in a measurable way.

\subsection{Extended Robustness Checks}

\begin{figure}[t]
    \centering
    \includegraphics[width=0.92\linewidth]{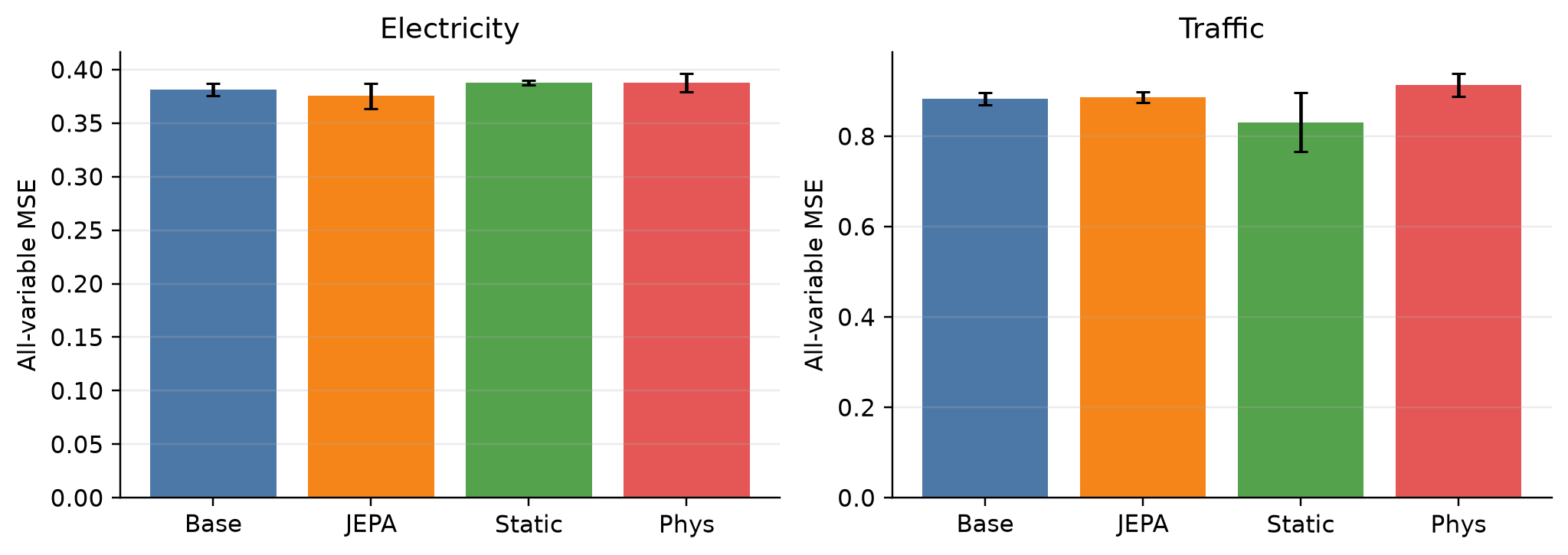}
    \caption{
    Compact H=24 repeated-seed robustness check.
    Because these runs use a faster setting than the main experiments, they are used to examine variance rather than to replace the main H=24 table.
    }
    \label{fig:compact-repeat}
\end{figure}

We additionally ran compact repeated-seed and baseline checks.
These runs use a faster setting with batch size 1024, 12 training epochs and patience 3, so they are reported as robustness checks rather than replacements for the main tables.
Table~\ref{tab:compact-repeat} reports H=24 seed-11/13 mean and standard deviation, while Figure~\ref{fig:compact-repeat} visualizes the aggregate MSE.

\begin{table}[t]
\centering
\caption{Compact H=24 repeated-seed robustness check. Values are mean $\pm$ standard deviation over seeds 11 and 13 under the compact training setting.}
\label{tab:compact-repeat}
\begin{tabular}{llrr}
\toprule
Dataset & Method & All MSE & Target MSE \\
\midrule
Electricity & Baseline & 0.381134 $\pm$ 0.005574 & 0.385227 $\pm$ 0.004184 \\
Electricity & JEPA & \textbf{0.375372 $\pm$ 0.011966} & \textbf{0.375935 $\pm$ 0.000035} \\
Electricity & Phys-JEPA-Static & 0.387680 $\pm$ 0.002376 & 0.404639 $\pm$ 0.013100 \\
Electricity & Phys-JEPA & 0.387938 $\pm$ 0.008497 & 0.383759 $\pm$ 0.015469 \\
Traffic & Baseline & 0.883816 $\pm$ 0.013545 & 0.658799 $\pm$ 0.024245 \\
Traffic & JEPA & 0.886540 $\pm$ 0.012059 & 0.643052 $\pm$ 0.042030 \\
Traffic & Phys-JEPA-Static & \textbf{0.831068 $\pm$ 0.065256} & \textbf{0.614187 $\pm$ 0.005367} \\
Traffic & Phys-JEPA & 0.914141 $\pm$ 0.025123 & 0.658675 $\pm$ 0.009897 \\
\bottomrule
\end{tabular}
\end{table}

The compact repeated-seed check is consistent with the descriptor-quality interpretation.
On Traffic, Phys-JEPA-Static is the most stable compact variant, improving aggregate MSE over the compact baseline.
Full Phys-JEPA is less robust in this shortened training setting, suggesting that the dynamic loss is more sensitive to optimization budget and loss weighting.
On Electricity, JEPA is strongest in the compact check, while full Phys-JEPA gives a lower target MSE than Phys-JEPA-Static.
We therefore treat these results as evidence about variance and sensitivity, not as the final repeated-seed benchmark.

\begin{table}[t]
\centering
\caption{Compact physical-alignment diagnostics at H=24. Correlation-based metrics are emphasized because $R^2$ can be unstable when weak-domain descriptor variance is very small.}
\label{tab:alignment}
\resizebox{\linewidth}{!}{%
\begin{tabular}{llrrr}
\toprule
Dataset & Method & Physical corr. & Transition corr. & Transition MSE \\
\midrule
Electricity & Phys-JEPA-Static & 0.370790 $\pm$ 0.019816 & 0.479497 $\pm$ 0.000441 & 0.782082 $\pm$ 0.071537 \\
Electricity & Phys-JEPA & 0.290310 $\pm$ 0.012174 & \textbf{0.557829 $\pm$ 0.000481} & 0.972061 $\pm$ 0.156036 \\
Traffic & Phys-JEPA-Static & \textbf{0.214373 $\pm$ 0.200664} & 0.331993 $\pm$ 0.020718 & 0.006746 $\pm$ 0.001496 \\
Traffic & Phys-JEPA & 0.041692 $\pm$ 0.085775 & \textbf{0.398316 $\pm$ 0.090912} & \textbf{0.006011 $\pm$ 0.000265} \\
\bottomrule
\end{tabular}
}
\end{table}

The alignment diagnostics provide direct evidence for the method design.
Full Phys-JEPA improves transition correlation over the static-only variant on both datasets, which supports the claim that the dynamic loss affects latent state transitions rather than only decoded outputs.
However, physical-state correlation is higher for the static-only variant, reflecting the expected trade-off between state alignment and transition alignment.

\begin{table}[t]
\centering
\caption{DLinear comparison under the compact training setting. DLinear was run with the same data split and context length. Phys-JEPA values are from the main/full horizon experiments, while DLinear values use the compact baseline implementation.}
\label{tab:dlinear}
\resizebox{\linewidth}{!}{%
\begin{tabular}{llrrrr}
\toprule
Dataset & Horizon & DLinear All MSE & Phys-JEPA All MSE & DLinear Target MSE & Phys-JEPA Target MSE \\
\midrule
Electricity & 24 & \textbf{0.185666} & 0.336799 & \textbf{0.325247} & 0.354296 \\
Electricity & 48 & \textbf{0.218166} & 0.394410 & \textbf{0.422672} & 0.564099 \\
Electricity & 96 & \textbf{0.219855} & 0.392119 & \textbf{0.397730} & 0.483457 \\
Electricity & 192 & \textbf{0.217662} & 0.391452 & \textbf{0.369930} & 0.501240 \\
Traffic & 24 & \textbf{0.673272} & 0.684625 & \textbf{0.309953} & 0.506928 \\
Traffic & 48 & 0.770982 & \textbf{0.726473} & \textbf{0.383148} & 0.526501 \\
Traffic & 96 & \textbf{0.716811} & 0.770301 & \textbf{0.335486} & 0.558793 \\
Traffic & 192 & \textbf{0.671076} & 0.773873 & \textbf{0.296076} & 0.595946 \\
\bottomrule
\end{tabular}
}
\end{table}

\begin{figure}[t]
    \centering
    \includegraphics[width=0.92\linewidth]{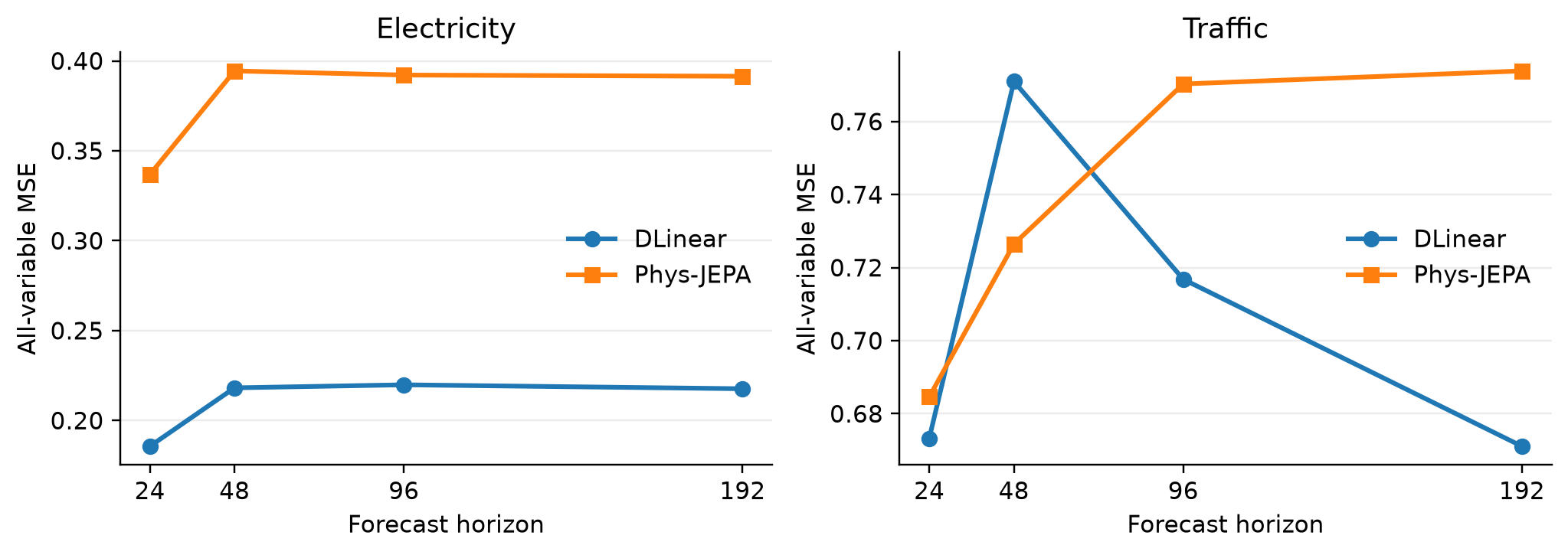}
    \caption{
    Compact DLinear comparison on the completed horizons.
    DLinear is a strong decomposition-based baseline; these results motivate a full strong-baseline suite before making broad claims against modern forecasting methods.
    }
    \label{fig:dlinear}
\end{figure}

The compact DLinear comparison is intentionally conservative for Phys-JEPA.
DLinear is strong on Electricity and on Traffic target-channel MSE, while Phys-JEPA remains competitive on Traffic aggregate MSE at H=24 and is better at H=48.
This reinforces the need to compare Phys-JEPA against strong forecasting backbones in future work, and it clarifies that the current contribution concerns where physical constraints are imposed rather than universal superiority over all forecasting architectures.

\begin{table}[t]
\centering
\caption{Compact H=24 hyperparameter sensitivity for the physical latent branch. The setting column reports $(d_{phy},\lambda_s,\lambda_d)$.}
\label{tab:compact-sweep}
\begin{tabular}{llrrrr}
\toprule
Dataset & Setting & All MSE & Target MSE & Physical corr. & Transition corr. \\
\midrule
Electricity & $(16,0.01,0.05)$ & \textbf{0.383562} & \textbf{0.371853} & \textbf{0.399356} & 0.539506 \\
Electricity & $(32,0.01,0.05)$ & 0.390624 & 0.404854 & 0.256608 & 0.531809 \\
Electricity & $(32,0.01,0.10)$ & 0.391671 & 0.404281 & 0.183070 & \textbf{0.555034} \\
Electricity & $(32,0.02,0.00)$ & 0.578443 & 0.712318 & -0.012120 & 0.420051 \\
Electricity & $(64,0.01,0.05)$ & 0.390314 & 0.387997 & 0.304435 & 0.531799 \\
Traffic & $(16,0.01,0.05)$ & 0.887293 & 0.654713 & 0.057651 & 0.397737 \\
Traffic & $(32,0.01,0.05)$ & 0.894238 & 0.618895 & 0.021550 & 0.409361 \\
Traffic & $(32,0.01,0.10)$ & 0.866117 & \textbf{0.610332} & -0.018554 & \textbf{0.431322} \\
Traffic & $(32,0.02,0.00)$ & \textbf{0.858845} & 0.637749 & -0.010105 & 0.267464 \\
Traffic & $(64,0.01,0.05)$ & 0.881501 & 0.632211 & \textbf{0.075056} & 0.338868 \\
\bottomrule
\end{tabular}
\end{table}

\begin{figure}[t]
    \centering
    \includegraphics[width=0.94\linewidth]{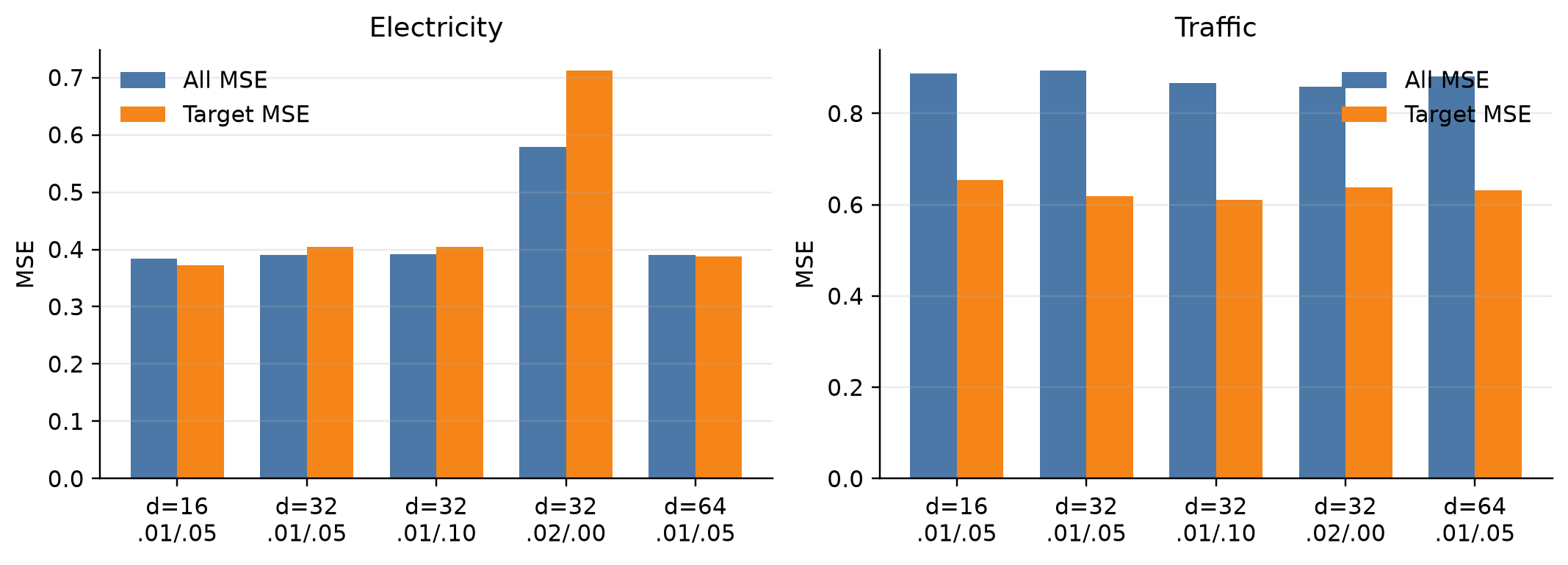}
    \caption{
    Compact H=24 sensitivity to physical latent dimension and static/dynamic loss weights.
    The sweep shows that the best accuracy and the strongest transition alignment do not always coincide, which is expected when weak-domain descriptors only partially observe the underlying dynamics.
    }
    \label{fig:compact-sweep}
\end{figure}

The compact hyperparameter sweep supports two design choices in Phys-JEPA.
First, increasing the dynamic weight from $0.05$ to $0.10$ improves transition correlation on both datasets, directly matching the intended role of $\mathcal{L}_{dyn}$.
Second, larger static weight without dynamic consistency is not automatically beneficial: on Electricity, the static-heavy setting substantially degrades accuracy and alignment, while on Traffic it improves aggregate MSE but weakens transition correlation.
This trade-off is consistent with the method section's argument that static state alignment and dynamic transition alignment regularize different aspects of the latent world model.

\subsection{Summary of Findings}

Across the experiments, three findings are most important.
First, JEPA is a reasonable temporal representation backbone but is not uniformly better than a supervised baseline across all horizons.
Second, output-level physics is weaker than latent physical regularization in the tested physical benchmarks.
Third, the best latent physical constraint depends on the quality of the physical descriptor and the prediction horizon: Traffic benefits consistently from combined static and dynamic consistency, while Electricity favors static consistency at shorter horizons and full Phys-JEPA at H=192.
These findings motivate the next experimental step: scaling the compact repeated-seed and strong-baseline checks to the full main-experiment setting and adding descriptor-quality ablations for Electricity and Traffic.

\section{Discussion}

The results suggest that physical consistency is more effective when it organizes latent predictive states than when it is applied only to decoded forecasts.
The Jena Climate case supports this interpretation because the physical branch is aligned with directly interpretable meteorological variables, and Phys-JEPA improves most thermodynamic and vapor-pressure channels.
The ablation study further shows that the two Phys-JEPA losses play different roles.
Static latent state consistency is robust on Electricity, while the full Static+Dynamic objective is strongest on Traffic.
This supports the hypothesis that physically meaningful latent states can improve multivariate forecasting by regularizing the internal world model, while also showing that transition consistency must be matched to the quality of the physical descriptor.

When the descriptor captures a coherent temporal process, dynamic consistency can help the model learn a more meaningful latent transition.
When the descriptor is coarse or aggregates many weakly related processes, the same transition loss can become over-constraining.
This motivates future work on stronger physical descriptors, horizon-dependent loss weighting and more expressive physics projectors.
More importantly, it motivates evaluating Phys-JEPA on datasets governed by stronger explicit physical constraints, where latent state consistency and latent transition consistency can be judged against clearer physical laws rather than weak domain descriptors.

The study has three main limitations.
First, only Jena Climate provides directly interpretable meteorological variables; Electricity and Traffic use weak-domain descriptors rather than exact governing equations, so the infrastructure results should be interpreted as evidence for latent physical regularization under practical descriptor construction rather than as proof of exact physical-law discovery.
Second, the benchmark tables are single-seed, so small differences should be treated cautiously until repeated-seed averages and standard deviations are available.
Third, the present comparison focuses on the supervised backbone, JEPA variants and physics ablations; stronger standalone forecasting baselines such as DLinear, PatchTST, TimesNet and iTransformer should be added under the same splits before making broad claims against modern LTSF methods.
Accordingly, we view this manuscript as an initial validation of the latent-physics formulation rather than a final benchmark study.

\section{Conclusion}

We introduced Phys-JEPA, a physics-informed latent world model for multivariate time-series forecasting.
By decomposing latent predictive states into physical and residual components, Phys-JEPA moves physical regularization from output space into latent predictive state space.
Experiments on Jena Climate, Electricity and Traffic show that latent physical constraints can improve forecasting when the descriptor captures meaningful state evolution.
On Jena Climate, Phys-JEPA improves both aggregate forecasting and the temperature target, supporting the value of latent physical alignment when variables have direct meteorological meaning.
Full Phys-JEPA is strongest on Traffic across all tested horizons, while Electricity shows a horizon-dependent pattern in which static consistency is best at shorter horizons and full static--dynamic consistency is best at H=192.
The ablation study further shows that output-level physics is weaker than latent physical regularization in this setting, supporting the central design choice of organizing the predictive representation itself.
Overall, these experiments indicate that the proposed direction warrants broader study, especially on problems with stronger physical constraints and clearer transition laws.
Future work should add repeated-seed evaluation, stronger modern forecasting baselines, descriptor-quality ablations and physically stricter benchmarks to clarify when dynamic latent physics is most beneficial.

\bibliographystyle{plain}
\bibliography{references}

\end{document}